\documentclass[10pt]{article}
\PassOptionsToPackage{dvipsnames,svgnames,x11names}{xcolor}
\usepackage[citestyle=authoryear]{colab}

\usepackage{amsmath,amsfonts,bm}

\def\eqref#1{equation~\ref{#1}}

\def\1{\bm{1}}

\DeclareMathAlphabet{\mathsfit}{\encodingdefault}{\sfdefault}{m}{sl}
\SetMathAlphabet{\mathsfit}{bold}{\encodingdefault}{\sfdefault}{bx}{n}

\usepackage{needspace}
\title{LTM: Latent Transition Based In-Context Memory for Vision-Language-Action Models}
\author{%
  Qixin Sun$^{1,2}$ \quad WenLong Gao$^{2}$ \quad Xiansheng Chen$^{2}$ \\[4pt]
  Zhongyuan Wang$^{2}$ \quad Si Liu$^{1,*}$ \quad Pengwei Wang$^{2,*}$ \\[6pt]
  {\normalsize $^{1}$Beihang University \quad $^{2}$BAAI} \\[3pt]
  {\small $^{*}$Corresponding authors}%
}
\labname{Colab}
\institution{}
\colabdate{October 2026}
\paperurl{}
\githuburl{}
\huggingfaceurl{}
\dataurl{}
\colabrunningtitle{LTM: Latent Transition Memory for Vision-Language-Action Models}
\hypersetup{pdftitle={LTM: Latent Transition Based In-Context Memory for Vision-Language-Action Models},pdfauthor={Qixin Sun, WenLong Gao, Xiansheng Chen, Zhongyuan Wang, Si Liu, Pengwei Wang}}

\begin{document}

\maketitle

\begin{colababstract}
\small
Robotic manipulation often requires remembering interactions that are no longer visible. We introduce \textbf{Latent Transition Memory (LTM)}, which stores history as compact latent transitions between consecutive observations. Instantiated on $\pi_{0.5}$ as \textbf{LTM-$\pi$}, our method uses a frozen latent action encoder, a lightweight adapter, and causal KV caching for incremental inference without memory-specific semantic annotations or external reasoning-model calls. With DreamDojo transitions, LTM-$\pi$ achieves 58.54\% success on RoboMME and 80.5\% on LIBERO-Mem, performing best among baselines without external reasoning models or memory-specific annotations. Ablations support the contribution of transition content beyond temporal cues and the efficiency of delayed memory admission. Scaling RoboMME demonstrations tenfold improves success from 57.67\% to 87.50\%, while a Cup Swap diagnostic highlights the role of process supervision in temporal composition. Under a 200-token memory budget, success remains 84.00\%, versus 87.50\% without compression. Finally, real-robot counting success improves from 11.25\% to 82.50\%. Together, these results demonstrate the effectiveness, scalability, and efficiency of latent transition memory, highlighting its capacity for complex temporal information, bounded-context operation, and real-world robotic control.
Our dataset, training code, and inference code have been publicly released at: \url{https://github.com/buaa-colalab/LTM-pi}.
\end{colababstract}

\section{Introduction}

\begin{wrapfigure}{r}{0.45\textwidth}
  \vspace{-6pt}
  \centering
  \includegraphics[width=\linewidth]{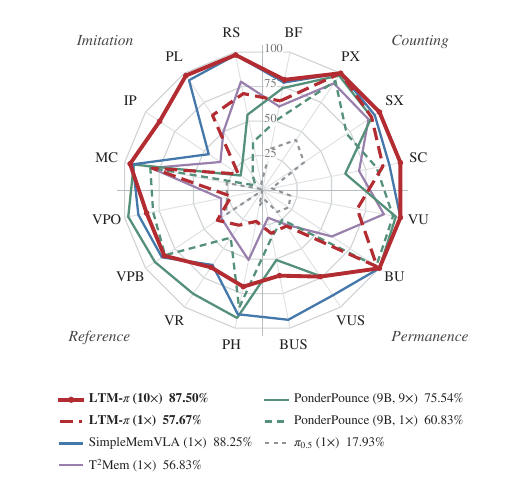}
  \captionsetup{font=footnotesize,skip=3pt,width=\linewidth}
  \caption{Success rates (\%) on the 16 RoboMME tasks, grouped by track.
  The legend shows data scales and overall success rates; scores follow Tables~\ref{tab:robomme_results} and~\ref{tab:robomme_scaling}.}
  \label{fig:teaser}
  \vspace{-6pt}
\end{wrapfigure}

Vision-language-action (VLA) models combine pretrained visual and linguistic representations with action generation to follow diverse manipulation instructions~\citep{brohan2023rt2,kim2024openvla,black2024pi0,physicalintelligence2025pi05}. Yet the current observation may omit information needed for the next action. A robot placing three objects into a bag must remember completed transfers; one tracking an occluded object must retain the effects of earlier interactions. Similar scenes can therefore require different actions depending on the preceding history, motivating explicit memory for robotic control~\citep{dai2026robomme,chung2026rethinking}.

Existing memory designs retain visual observations, summarize experience in language, or update a recurrent state~\citep{shi2025memoryvla,qu2026lamemvla,torne2026mem,cherepanov2026muvla}. Dense visual histories consume substantial context, whereas sparse sampling can omit intermediate interactions. Language summaries abstract away visual detail but may miss fine-grained motion. Recurrent representations support compact updates but must continually decide what to preserve in a bounded state. 

These trade-offs raise a representational question: \textbf{\emph{what should a VLA remember about its past interactions?}}

For many manipulation tasks, the relevant history is the sequence of changes induced by robot--environment interactions. Latent action models encode an observation pair into a compact code from which the later observation can be reconstructed conditioned on the earlier one~\citep{bruce2024genie,ye2024lapa}. This objective encourages the code to capture temporal changes. While latent actions have been used for policy pretraining and world-model control~\citep{ye2024lapa,bu2025univla,gao2026dreamdojo}, we study their use as an ordered episodic memory for action generation.

We introduce \textbf{Latent Transition Memory (LTM)}, which encodes consecutive observations with a frozen latent action model and projects each code into one memory token. The policy receives a delayed transition prefix alongside the instruction and current observation. Each code depends only on an adjacent observation pair, allowing encoding to overlap action execution. We instantiate LTM on $\pi_{0.5}$ as \textbf{LTM-$\pi$}, with a lightweight adapter and causal attention that permits persistent KV caching. Sparse initial-frame anchors provide visual grounding for video-conditioned tasks. The policy and adapter learn from robot demonstrations without memory-specific semantic annotations. At inference, the frozen encoder supplies transition codes directly to the policy, without an external reasoning model or intermediate language generation.

We evaluate LTM-$\pi$ with two frozen latent encoders, CD-LAM and DreamDojo, on RoboMME and LIBERO-Mem. With the original benchmark data, the DreamDojo variant reaches 58.54\% average success on RoboMME, compared with 17.93\% for the reported memory-free $\pi_{0.5}$ baseline, and 80.5\% on LIBERO-Mem, compared with 42.5\% for the strongest baseline included in our comparison. We then isolate the effects of memory content and inference lag. Replacing observation-dependent latent transitions with fixed random representations reduces RoboMME success from 57.67\% to 44.50\%, supporting the contribution of transition-specific content beyond temporal-progress cues. Using a five-step lag reduces estimated latent-encoder blocking from 114.25\,ms to 9.47\,ms relative to zero lag, with a small change in success (57.88\% to 57.67\%).

We next test how LTM benefits from broader demonstration coverage. We expand RoboMME from 100 to 1,000 episodes per task family and increase training from 40K to 120K steps. With the same encoder and policy architecture, success rises from 57.67\% to 87.50\% (Figure~\ref{fig:teaser}). All four tracks improve; Reference, the weakest base-data track, gains 47.50 percentage points. Extending training to 120K steps on the original data alone yields no appreciable gain, supporting the benefit of broader demonstration coverage. LTM-$\pi$ exceeds the reported PonderPounce result with $9\times$ data by 11.96 points overall, under different data budgets and supervision assumptions.

Scaling does not fully resolve the challenge of composing local transitions into persistent object states: VideoUnmaskSwap and ButtonUnmaskSwap reach 74\% and 62\%, versus 100\% for their non-swap counterparts. In a Cup Swap diagnostic, process supervision achieves 97.93\% final-state recovery from latent transitions, compared with 2.40\% under final-state-only supervision. These results demonstrate that latent transitions preserve recoverable multi-step interaction information and highlight the importance of intermediate supervision, motivating language-guided reasoning and process supervision to address remaining temporal-composition challenges.

To bound execution-memory growth, we merge the most similar adjacent projected tokens under the same quota during training and inference. On RoboMME-10x, a 200-token budget retains 84.00\% success versus 87.50\% uncompressed and reduces mean transition-token count by at least 38.26\% on profiled training queries.

Finally, we evaluate LTM-$\pi$ on a Galbot G1 robot in a real-world repeated pick-and-place counting task. LTM-$\pi$ improves average success from 11.25\% for the memory-free $\pi_{0.5}$ baseline to 82.50\%, demonstrating that transition memory also helps track completed interactions during physical manipulation.

\begin{samepage}
Our contributions are threefold:
\begin{itemize}
  \item We propose \textbf{Latent Transition Memory (LTM)}, a memory representation that uses compact latent transitions between consecutive observations as historical context instead of retaining the full visual history. This focuses memory on interaction-induced changes and enables the policy to use past experience without external reasoning-model calls or memory-specific semantic annotations.
  \item We develop \textbf{LTM-$\pi$}, integrating transition memory into $\pi_{0.5}$'s inference context through a lightweight adapter. Causal attention, persistent KV caching, and asynchronous transition encoding support efficient online control. Experiments on RoboMME and LIBERO-Mem demonstrate strong memory-dependent manipulation performance, with a measured policy inference time of 66.71\,ms on an H100.
  \item We characterize \textbf{transition-memory information, data scaling, and context-budget trade-offs}. Ablations distinguish transition content from temporal cues; tenfold RoboMME data expansion raises success from 57.67\% to 87.50\%; process-supervised Cup Swap decoding recovers final states with 97.93\% accuracy; and a 200-token budget retains 84.00\% success. Finally, real-robot counting experiments improve success from 11.25\% to 82.50\%.
\end{itemize}
\end{samepage}

\section{Related Work}

\paragraph{Vision-language-action models.}
VLA models transfer pretrained visual and linguistic representations to robotic control. RT-2~\citep{brohan2023rt2} predicts tokenized actions, and OpenVLA~\citep{kim2024openvla} provides an open-source generalist policy. $\pi_0$~\citep{black2024pi0} introduces flow matching for continuous action generation; $\pi_{0.5}$~\citep{physicalintelligence2025pi05} and LingBot-VLA~\citep{wu2026lingbotvla} extend generalization through heterogeneous co-training and multi-embodiment data. These advances do not remove partial observability: the appropriate action may depend on events absent from the current observation. RoboMME~\citep{dai2026robomme} and LIBERO-Mem~\citep{chung2026rethinking} evaluate this need for execution memory.

\paragraph{Memory mechanisms for VLAs.}
RoboFlamingo~\citep{li2023roboflamingo} models history through a sequential policy head; HAMLET~\citep{koo2025hamlet} and $\mu$VLA~\citep{cherepanov2026muvla} use learned memory tokens. MemoryVLA~\citep{shi2025memoryvla} and LaMem-VLA~\citep{qu2026lamemvla} retrieve evidence from explicit memory banks. SAM2Act+~\citep{fang2025sam2act} extends a multi-view manipulation policy with a spatial memory bank, encoder, and memory attention. Object-centric approaches retain a different unit of history: SlotVLA~\citep{hanyu2025slotvla} models object slots and their relations, while E-SlotSSM~\citep{chung2026rethinking} studies the progression of object-centric memory states. GMP~\citep{gao2026gatedmemorypolicy} learns when to activate cross-attention over cached image and action tokens. RoboMME also compares symbolic, sampled-frame, and recurrent memories with several policy interfaces~\citep{dai2026robomme}. LTM instead retains an ordered sequence of changes between observations, rather than appearance tokens or a recurrent summary of the episode. Temporal prediction also informs memory-based control: HiF-VLA~\citep{lin2025hifvla} incorporates motion representations, MemoryVLA++~\citep{shi2026memoryvlapp} combines retrieved history with imagined futures, and MemoryVAM~\citep{jiang2026memoryvam} injects compressed visual memory into both video prediction and action decoding.

MEM~\citep{torne2026mem} combines recent video with long-term textual memory, HiMe~\citep{ji2026hime} separates working memory from long-term planning, and WeaveLA~\citep{zhu2026weavela} passes latent summaries of completed segments to subsequent subtasks. MemER~\citep{sridhar2025memer} retrieves keyframes for a high-level vision-language model that generates instructions for a low-level policy. PonderPounce~\citep{choi2026ponderpounce} uses a separate pretrained multimodal context model to asynchronously supply continuous cognition tokens to its action model. Its RoboMME training recipe includes transition, subgoal, and demonstration-reasoning supervision. In 2AM~\citep{hu20262am}, an agent maintains episodic memory and guides an episodically stateless action model through subtask language and optional 2D spatial hints. These designs delegate historical interpretation to a higher-level model or interface. LTM instead exposes observed transitions directly to the action policy, without invoking a separate reasoning model or generating intermediate language during control.

SimpleMemVLA~\citep{yin2026simplememvla} retains timestamped video in its VLM's native context and reuses the prefix cache through streaming prefill. Its action head receives representations of generated subtask text, with subtask annotations produced offline by a cloud VLM. This is a training-time annotation dependency, rather than a requirement for cloud-model calls during control. T$^2$Mem~\citep{liu2026t2mem} places a fast-weight memory between the VLM and action expert and updates it through self-supervised test-time gradients. Like LTM, it learns from action demonstrations without an external reasoning model or memory-specific annotations. Our distinction is therefore the memory representation and update rule: LTM appends frozen pairwise transition codes to a persistent sequence, without test-time gradient updates. Its causal attention and delayed prefix support incremental inference while preserving access to the current observation.

\paragraph{Latent action models.}
LAMs infer compact transition codes from observation pairs, typically using reconstruction without action labels~\citep{bruce2024genie,ye2024lapa}. Genie~\citep{bruce2024genie} uses discrete codes for controllable video generation, LAPA~\citep{ye2024lapa} uses them for policy pretraining, and CLAM~\citep{liang2025clam} and UniVLA~\citep{bu2025univla} extend latent actions to continuous and cross-embodiment control. DreamDojo~\citep{gao2026dreamdojo} uses latent actions for world-model pretraining; CD-LAM~\citep{wei2026cdlam} reduces visual confounding in such codes. We use pretrained LAMs only to encode observed transitions into episodic memory; the deployed policy does not need to reconstruct past frames or generate future videos.

\Needspace{10\baselineskip}
\section{Method}
\begin{figure}[ht]
  \centering
  \includegraphics[width=0.9\linewidth]{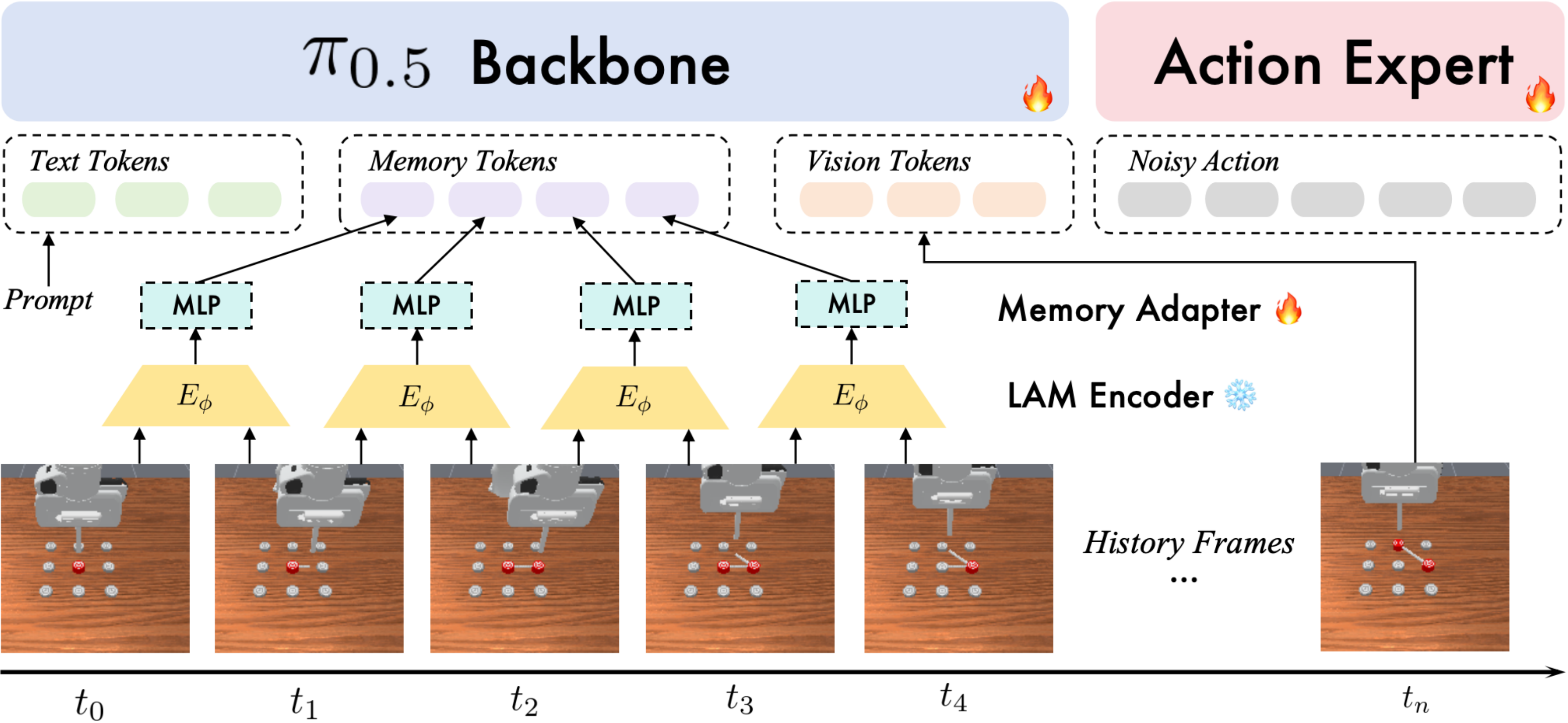}
  \caption{LTM-$\pi$ encodes consecutive observations with a frozen latent action model and maps each transition to a memory token. The trainable policy conditions on the instruction, transition history, and current observation. The schematic omits the fixed lag and the visual anchors used for video demonstrations.}
  \label{fig:main}
\end{figure}
\subsection{Latent-Transition-Based In-Context Memory for VLAs}

We represent interaction history using compact latent
transitions between consecutive observations.
A latent action model (LAM) encodes each transition from $o_{i-1}$ to
$o_i$ into a latent code $z_i$, from which a decoder reconstructs $o_i$
conditioned on $o_{i-1}$:
\[
z_i = E_\phi(o_{i-1},o_i), \qquad
\hat{o}_i = D_\psi(o_{i-1},z_i), \qquad
\mathcal{L}_{\mathrm{LAM}} = d(\hat{o}_i,o_i),
\]
where $E_\phi$ and $D_\psi$ are the transition encoder and decoder,
respectively, and $d$ is a reconstruction distance. Conditioning the
decoder on the preceding observation encourages the compact latent
bottleneck to capture temporal changes rather than redundant appearance
information.

Two properties make these representations useful for VLA memory:
\textbf{(i) Compactness.} Each transition is represented by a
low-dimensional vector, reducing the context budget required to retain
interaction history.
\textbf{(ii) Locality.} Each code depends only on two consecutive
observations, allowing incremental memory construction without
re-encoding earlier frames.

Let $\mathbf A_t$ denote an action chunk, $s_t$ the robot state, and
$\ell$ the language instruction. A current-observation policy predicts
$\mathbf A_t \sim \pi_\theta(\cdot \mid o_t,s_t,\ell)$.
Existing memory mechanisms augment this policy in different ways:
frame-based memory uses sampled observations,
$\mathbf A_t \sim
\pi_\theta(\cdot \mid o_t,s_t,\ell,
\{o_i\}_{i \in \mathcal I_t})$,
where $\mathcal I_t \subseteq \{0,\ldots,t-1\}$;
language-based memory uses a summary
$m_t^{\mathrm{lang}} = f_{\mathrm{LLM}}(o_{<t})$,
giving $\mathbf A_t \sim
\pi_\theta(\cdot \mid o_t,s_t,\ell,m_t^{\mathrm{lang}})$;
and recurrent memory updates
$h_t = f_{\mathrm{rec}}(h_{t-1},o_t)$,
giving $\mathbf A_t \sim
\pi_\theta(\cdot \mid o_t,s_t,\ell,h_t)$.

Our approach instead conditions the policy on an ordered sequence of
latent transitions:
\[
\mathcal M_t^{\mathrm{LAM}}
= (z_i)_{1 \leq i \leq \tau_t},
\qquad
\mathbf A_t \sim
\pi_\theta(\cdot \mid o_t,s_t,\ell,\mathcal M_t^{\mathrm{LAM}}).
\]
Here, $\tau_t=\max(0,t-k)$ excludes the latest $k$ transitions,
while $o_t$ remains the current visual input. This representation retains transition-level information
without dense historical RGB tokens or recurrent compression; video-conditioned tasks additionally use the sparse anchors described below.

\subsection{LTM-$\pi$}

We instantiate this memory mechanism on $\pi_{0.5}$ and refer to the
resulting model as \textbf{LTM-}\(\boldsymbol{\pi}\).
The policy and memory projector are trained on demonstration data
using a frozen LAM encoder, without requiring language summaries or
intermediate semantic supervision. The architecture of \textbf{LTM-}\(\boldsymbol{\pi}\) is shown in Figure~\ref{fig:main}.

\paragraph{Training Phase.}
For each training sample, we use a fixed lag $k$, shared with inference, and set
$\tau_t=\max(0,t-k)$. Frozen DreamDojo~\citep{gao2026dreamdojo}
or CD-LAM~\citep{wei2026cdlam} encodes the retained observation pairs
into 32-dimensional transition vectors:
\[
z_i = E_{\phi}(o_{i-1},o_i) \in \mathbb R^{32},
\qquad
\tilde z_i = \mathtt{stopgrad} \left(
\frac{z_i-\mu_{\mathcal D}}{\sigma_{\mathcal D}}\right),
\quad 1 \leq i \leq \tau_t,
\]
where $E_{\phi}$ denotes the LAM encoder, and
$\mu_{\mathcal D}$ and $\sigma_{\mathcal D}$ are dataset-level
normalization statistics.

We use a lightweight memory adapter to align latent actions with the semantic space of the VLM.
A two-layer MLP projector maps each normalized transition into a
memory token in the 2048-dimensional hidden space of the
PaliGemma-3B backbone of $\pi_{0.5}$:
\[
m_i 
= \mathrm{LN}\!\left(
W_2\,\mathrm{GELU}(W_1\tilde z_i+b_1)+b_2
\right),
\qquad
W_1 \in \mathbb R^{512 \times 32},
\quad
W_2 \in \mathbb R^{2048 \times 512}.
\]

We insert these tokens between the language and current visual tokens:
\[
X_t =
\big[
X^{\mathrm{lang}}(\ell);
m_1,\ldots,m_{\tau_t};
X^{\mathrm{vis}}(o_t)
\big].
\]
The resulting context conditions the action expert. We apply causal attention within the language--memory
prefix and allow the current visual tokens to attend to this prefix.
Each prefix token therefore depends only on itself and preceding
prefix tokens, enabling its KV representations to be reused when
new memory tokens and visual inputs arrive.

\paragraph{Inference Phase.}
Algorithm~\ref{alg:LTM_pi_inference} summarizes online inference. At decision step $t$, the policy consumes transitions only through $\tau_t=\max(0,t-k)$. Each transition is encoded once, asynchronously after its observation pair becomes available, and stored in a latent buffer; the lag delays only when it is admitted to policy memory. Thus, a transition has up to $k$ observation steps of computation slack before it is required. At each decision, newly eligible latents are appended to the persistent language--memory KV cache, after which $o_t$ is processed and an action chunk is generated. This overlaps LAM computation with action execution and reduces critical-path blocking.

For $n$ cached prefix tokens and $b$ newly processed memory and visual
tokens, caching reduces the per-chunk backbone self-attention cost
from $\mathcal O((n+b)^2)$ to $\mathcal O(nb+b^2)$.
For fixed $b$, this cost is linear in the cached history length. In the append-only variant, the cache itself also grows linearly with the number of retained transitions. Section~\ref{sec:token_compression} evaluates an optional compression rule that bounds execution-memory length; merging existing tokens requires refreshing the affected cache entries.

\begin{algorithm}[htbp]
\caption{Online Inference of LTM-$\pi$}
\label{alg:LTM_pi_inference}
\begin{algorithmic}[1]
\Require Instruction $\ell$, frozen encoder $E_{\mathrm{LAM}}$, projector $P$, policy $\pi_\theta$, statistics $(\mu_{\mathcal D},\sigma_{\mathcal D})$, lag $k$, replanning interval $h$
\State $j,c\gets0$; $\mathcal C\gets\operatorname{Prefill}(X^{\mathrm{lang}}(\ell))$
\State Receive $o_0$
\While{task is not terminated}
  \State $t\gets j$; $\tau_t\gets\max(0,t-k)$; receive state $s_t$
  \State Wait only for unfinished $z_i$ with $c<i\leq\tau_t$
  \For{$i=c+1,\ldots,\tau_t$}
    \State $m_i\gets P((z_i-\mu_{\mathcal D})/\sigma_{\mathcal D})$; $\mathcal C\gets\operatorname{AppendKV}(\mathcal C,m_i)$; $c\gets i$
  \EndFor
  \State $\mathcal C_t\gets\operatorname{AppendKV}(\mathcal C,X^{\mathrm{vis}}(o_t))$
  \State $\mathbf A_t\sim\pi_\theta(\cdot\mid\mathcal C_t,s_t)$
  \While{executing the first $h$ actions of $\mathbf A_t$}
    \State Receive $o_{j+1}$; $j\gets j+1$
    \State LaunchAsync $z_j\gets E_{\mathrm{LAM}}(o_{j-1},o_j)$
    \Comment{store in latent buffer}
  \EndWhile
  \State Discard temporary visual KV entries; retain $\mathcal C$
\EndWhile
\end{algorithmic}
\vspace{4pt}
\noindent The algorithm shows the execution-memory path without a video demonstration. For video-conditioned tasks, the fixed demonstration tokens and initial-frame anchors are included in the persistent prefix as described in the method. We use $h=10$ on RoboMME and $h=5$ on LIBERO-Mem; $s_t$ is omitted on RoboMME.
\end{algorithm}

\paragraph{Visual Anchors for Video Demonstrations.}
For tasks conditioned on video demonstrations, the initial scene layout may differ between the demonstration and execution environments. To improve visual--semantic alignment, we complement latent transitions with sparse visual features from both sequences, inspired by frame-based memory sampling. A minimal choice uses the first frame of each sequence as a visual anchor:
\[
X_t = \big[X^{\mathrm{lang}}(\ell); X^{\mathrm{vis}}(o_0^{\mathrm{demo}}); m_1^{\mathrm{demo}},\ldots,m_{T_{\mathrm{demo}}}^{\mathrm{demo}}; X^{\mathrm{vis}}(o_0); m_1,\ldots,m_{\tau_t}; X^{\mathrm{vis}}(o_t)\big].
\]
Here, $o_0^{\mathrm{demo}}$ and $o_0$ are the initial demonstration and execution observations, respectively, and $m_i^{\mathrm{demo}}$ encodes the transition from $o_{i-1}^{\mathrm{demo}}$ to $o_i^{\mathrm{demo}}$. We add distinct learnable type embeddings to the demonstration and execution segments to distinguish their roles; these embeddings are omitted from the formula for clarity. The fixed visual anchors and demonstration tokens are retained in the persistent KV cache alongside the language and execution memory tokens.

\Needspace{10\baselineskip}
\section{Experiment}
Our experiments examine the effectiveness, information content, and memory cost of LTM-$\pi$. In simulation (Section~\ref{sec:simulation}), we introduce the benchmarks and evaluation protocol (Section~\ref{sec:sim_settings}), report results on RoboMME and LIBERO-Mem (Section~\ref{sec:sim_results}), and isolate the effects of memory components and lag (Section~\ref{sec:ablations}). We then examine demonstration scaling (Section~\ref{sec:data_scaling}), probe whether latent transitions encode swap information (Section~\ref{sec:swap_diagnostic}), and test manipulation under a fixed token budget (Section~\ref{sec:token_compression}). Finally, Section~\ref{sec:realworld_eval} presents the real-robot setup and repeated pick-and-place evaluation.

\subsection{Evaluation in Simulation}
\label{sec:simulation}
\subsubsection{Settings}
\label{sec:sim_settings}
\paragraph{Benchmarks.}
We evaluate memory-dependent robotic control on two complementary benchmarks, RoboMME~\citep{dai2026robomme} and LIBERO-Mem~\citep{chung2026rethinking}. RoboMME contains 16 history-dependent manipulation tasks organized into four suites: Counting, Permanence, Reference, and Imitation, targeting temporal, spatial, object, and procedural memory, respectively. These tasks require the policy to retain information from earlier observations or demonstrations that may no longer be recoverable from the current observation. LIBERO-Mem further evaluates long-horizon memory under object-level partial observability. It contains 10 tasks spanning four object-centric memory dimensions: Object Motion (OM), Object Sequence (OS), Object Relations (OR), and Object Occlusion (OO). The benchmark introduces temporally dependent subgoals and visually ambiguous objects, requiring the policy to track object identities, interactions, and state changes over trajectories of up to several hundred frames.

\paragraph{Training and Evaluation Configuration.}
We train the model on 8 H100 GPUs with a batch size of 128, using 40K steps on the original RoboMME data and 30K steps on LIBERO-Mem. The RoboMME-10x experiment uses 120K steps (Section~\ref{sec:data_scaling}). The learning rate decays from $2.5\times10^{-5}$ to $2.5\times10^{-6}$, with 1K warm-up steps. Unless otherwise specified, we use a fixed lag $k=5$ during training and inference. For the RoboMME data-scaling and token-compression experiments, we use $k=10$. The policy predicts action chunks of 50 steps on RoboMME and 10 steps on LIBERO-Mem, and replans after executing the first 10 and 5 steps, respectively. We do not use proprioceptive inputs on RoboMME. Following the official evaluation protocols, we evaluate each task on 20 layouts in LIBERO-Mem and over 50 rollouts in RoboMME. For the base-data RoboMME results in Table~\ref{tab:robomme_results}, we evaluate with three random seeds and report mean success rates. The Full and $k=5$ rows in the ablation tables reproduce these reference scores; the remaining ablations use a single seed.

\paragraph{Baselines.}
For RoboMME, we compare against DP-Unet, OpenVLA-OFT, $\pi_{0.5}$ with and without past actions, and its symbolic, perceptual, and recurrent memory variants~\citep{dai2026robomme}. We also include MemoryVLA~\citep{shi2025memoryvla}, SAM2Act+~\citep{fang2025sam2act}, and MemER~\citep{sridhar2025memer} as reported in RoboMME, alongside WeaveLA~\citep{zhu2026weavela}, PonderPounce (9B)~\citep{choi2026ponderpounce}, SimpleMemVLA~\citep{yin2026simplememvla}, and T$^2$Mem~\citep{liu2026t2mem}. To make the resource assumptions explicit, Table~\ref{tab:robomme_results} separates methods that use neither an external reasoning model nor memory-specific annotations from methods that use either resource. All baseline results are taken directly from published reports, with oracle-assisted methods and human performance listed separately for reference. Table~\ref{tab:robomme_results} uses the original data scale; Table~\ref{tab:robomme_scaling} additionally includes PonderPounce with $9\times$ data. These comparisons retain the reported methods' different training and supervision resources.
For LIBERO-Mem, we compare against $\pi_{0.5}$, SlotVLA~\citep{hanyu2025slotvla}, and Naive E-SlotSSM as reported by \citet{chung2026rethinking}, MemoryVLA~\citep{shi2025memoryvla}, GMP~\citep{gao2026gatedmemorypolicy}, MemoryVAM with UNet and DiT backbones~\citep{jiang2026memoryvam}, and 2AM together with its $\pi_0$ reimplementation~\citep{hu20262am}. 

\subsubsection{Main Results}
\label{sec:sim_results}
\paragraph{Results on RoboMME.}

\begin{table*}[h]
\centering
\fontsize{8}{9}\selectfont
\setlength{\tabcolsep}{3pt}
\renewcommand{\arraystretch}{1.0}
\caption{Success rates (\%, $\uparrow$) on all 16 RoboMME tasks. Deployable methods are divided into two blocks according to whether they avoid both external reasoning-model calls and memory-specific annotations. Within each block, bold and underline indicate the best and second-best results, respectively. Oracle and human references are excluded from ranking.}
\label{tab:robomme_results}
\resizebox{0.95\textwidth}{!}{%
\begin{tabular}{@{}l@{\hspace{8pt}}*{4}{r}@{\hspace{9pt}}*{4}{r}@{\hspace{9pt}}*{4}{r}@{\hspace{9pt}}*{4}{r}@{\hspace{9pt}}r@{}}
\toprule
Method & \multicolumn{4}{c}{Counting} & \multicolumn{4}{c}{Permanence} & \multicolumn{4}{c}{Reference} & \multicolumn{4}{c}{Imitation} & Avg. \\
\cmidrule(lr){2-5}\cmidrule(lr){6-9}\cmidrule(lr){10-13}\cmidrule(lr){14-17}
 & BF & PX & SX & SC & VU & BU & VUS & BUS & PH & VR & VPB & VPO & MC & IP & PL & RS & \\
\midrule
\multicolumn{18}{l}{\textit{Privileged-information and human references (not ranked)}} \\
SimpleSG+Oracle & 85.78 & 99.78 & 100.00 & 44.67 & 33.11 & 22.00 & 15.56 & 15.56 & 44.00 & 27.78 & 31.33 & 26.00 & 87.33 & 10.00 & 95.33 & 55.11 & 49.58 \\
GroundSG+Oracle & 85.78 & 100.00 & 100.00 & 49.67 & 98.78 & 95.00 & 99.22 & 80.22 & 83.33 & 97.33 & 100.00 & 100.00 & 87.78 & 15.56 & 97.00 & 55.56 & 84.08 \\
WeaveLA (oracle) & 28.00 & 34.00 & 50.00 & 12.00 & 24.00 & 28.00 & 22.00 & 4.00 & 12.00 & 2.00 & 22.00 & 14.00 & 16.00 & 2.00 & 8.00 & 12.00 & 18.12 \\
Human & 96.00 & 100.00 & 80.00 & 78.00 & 90.00 & 92.00 & 92.00 & 90.00 & 92.00 & 92.00 & 98.00 & 90.00 & 90.00 & 98.00 & 84.00 & 86.00 & 90.50 \\
\midrule
\multicolumn{18}{l}{\textit{External reasoning model and/or memory-specific annotations}} \\
SimpleSG+Gemini & 46.00 & 63.00 & 45.00 & 2.00 & 29.00 & 9.00 & 14.00 & 2.00 & 20.00 & 15.00 & 26.00 & 29.00 & 61.00 & 4.00 & 7.00 & 0.00 & 23.25 \\
SimpleSG+QwenVL & \underline{77.56} & \underline{95.33} & 5.11 & 0.44 & 34.22 & 19.33 & 15.33 & 9.56 & 17.11 & 25.33 & 33.33 & 25.11 & 82.00 & 3.78 & 12.67 & 7.78 & 29.00 \\
GroundSG+Gemini & 26.00 & 18.00 & 4.00 & 3.00 & 36.00 & 14.00 & 13.00 & 0.00 & 9.00 & 17.00 & 12.00 & 7.00 & 17.00 & 0.00 & 7.00 & 2.00 & 11.56 \\
GroundSG+QwenVL & 52.00 & 92.67 & 7.33 & 0.00 & 88.67 & 24.00 & 30.67 & 14.00 & 15.11 & 25.33 & 54.00 & 31.78 & 71.56 & 3.33 & 6.67 & 6.00 & 32.70 \\
MemER & 56.67 & 79.33 & 59.33 & 0.00 & 81.33 & 72.00 & \underline{38.00} & 21.33 & 70.67 & 25.33 & 30.00 & 26.00 & \underline{82.67} & \underline{6.67} & \underline{16.67} & 12.00 & 42.38 \\
WeaveLA & 26.00 & 44.00 & 46.00 & 2.00 & 28.00 & 14.00 & 14.00 & 4.00 & 10.00 & 4.00 & 26.00 & 18.00 & 22.00 & 4.00 & 8.00 & 8.00 & 17.38 \\
PonderPounce (9B) & 50.67 & 93.33 & \underline{72.00} & \underline{82.67} & \underline{94.67} & \underline{97.33} & 25.33 & \underline{34.00} & \underline{85.33} & \underline{40.67} & \underline{83.33} & \underline{79.33} & 81.33 & 6.00 & 12.67 & \underline{34.67} & \underline{60.83} \\
SimpleMemVLA & \textbf{78.00} & \textbf{100.00} & \textbf{96.00} & \textbf{92.00} & \textbf{100.00} & \textbf{100.00} & \textbf{90.00} & \textbf{94.00} & \textbf{90.00} & \textbf{64.00} & \textbf{86.00} & \textbf{90.00} & \textbf{94.00} & \textbf{46.00} & \textbf{94.00} & \textbf{98.00} & \textbf{88.25} \\
\midrule
\multicolumn{18}{l}{\textit{No external reasoning model; no memory-specific annotations}} \\
DP-Unet & 32.00 & 2.67 & 6.67 & 0.00 & 16.00 & 4.00 & 4.00 & 0.00 & 0.00 & 0.00 & 22.00 & 5.67 & 8.67 & 0.00 & 2.00 & 10.00 & 7.11 \\
OpenVLA-OFT & 8.00 & 8.00 & 2.00 & 2.00 & 16.00 & 0.00 & 18.00 & 0.00 & 10.00 & 0.00 & 28.00 & 20.00 & 18.00 & 4.00 & 6.00 & 8.00 & 9.25 \\
$\pi_{0.5}$ & 30.00 & 42.89 & 35.56 & 6.67 & 20.44 & 22.22 & 18.67 & 6.67 & 11.33 & 0.44 & 31.11 & 25.78 & 26.00 & 1.56 & 2.89 & 4.67 & 17.93 \\
$\pi_{0.5}$ w/ past actions & 26.67 & 58.33 & 26.67 & 4.67 & 30.67 & 23.67 & 20.67 & 16.00 & 12.33 & 8.67 & 24.00 & 18.67 & 34.00 & 1.00 & 4.00 & 5.67 & 19.73 \\
TokenDrop+Context & 48.67 & 85.11 & \underline{94.67} & 3.11 & 33.78 & 31.56 & 26.22 & 16.00 & 20.67 & 17.78 & 31.11 & 25.33 & 81.33 & 4.00 & 12.67 & 20.00 & 34.50 \\
TokenDrop+Modul & 34.44 & 83.56 & 86.00 & 5.33 & 28.22 & 29.33 & 28.44 & 21.33 & 21.33 & 22.00 & \underline{59.56} & \textbf{36.00} & 62.00 & 7.11 & 32.44 & 51.56 & 38.04 \\
TokenDrop+Expert & 54.22 & 87.56 & 91.78 & 4.22 & 26.67 & 30.44 & 18.44 & 18.89 & 19.33 & 20.89 & 36.44 & 24.67 & \textbf{87.56} & 2.22 & 16.22 & 18.22 & 34.86 \\
FrameSamp+Context & 41.22 & 72.00 & 73.67 & 13.67 & 26.89 & 30.22 & 20.89 & 15.22 & 17.67 & 15.22 & 30.00 & 20.89 & 77.22 & 1.22 & 15.22 & 19.67 & 30.68 \\
FrameSamp+Modul & 39.56 & 87.33 & 92.00 & 42.00 & 32.67 & 25.11 & 24.44 & 18.22 & 22.89 & 30.44 & \textbf{60.00} & \underline{32.00} & 77.78 & 7.56 & 53.56 & 66.67 & 44.51 \\
FrameSamp+Expert & 57.33 & 86.22 & \underline{94.67} & 28.89 & 31.78 & 25.78 & 22.89 & 20.22 & 19.11 & 23.11 & 30.00 & 24.22 & \underline{83.11} & 2.00 & 13.56 & 17.11 & 36.25 \\
TTT+Context & 35.56 & 62.89 & 42.44 & 3.33 & 29.78 & 22.89 & 18.44 & 14.44 & 20.44 & 13.11 & 34.22 & 20.22 & 32.44 & 1.11 & 1.56 & 3.56 & 22.28 \\
TTT+Modul & 34.22 & 65.11 & 36.67 & 2.11 & 27.22 & 22.11 & 25.22 & 14.11 & 14.56 & 12.11 & 32.67 & 22.33 & 31.22 & 1.11 & 3.56 & 7.00 & 21.96 \\
TTT+Expert & 34.89 & 63.78 & 41.33 & 4.00 & 31.78 & 22.44 & 19.56 & 18.00 & 12.22 & 9.56 & 34.00 & 22.89 & 33.56 & 0.89 & 3.11 & 5.56 & 22.35 \\
RMT+Context & 32.44 & 56.89 & 33.56 & 5.78 & 31.33 & 10.89 & 17.33 & 2.00 & 14.00 & 3.78 & 32.00 & 29.11 & 25.78 & 2.00 & 5.56 & 8.89 & 19.46 \\
RMT+Modul & 33.33 & 60.78 & 37.78 & 4.67 & 31.11 & 11.78 & 17.78 & 2.44 & 17.11 & 4.22 & 32.00 & 31.11 & 24.67 & 2.21 & 3.78 & 8.00 & 20.17 \\
RMT+Expert & 35.78 & 60.22 & 36.00 & 5.56 & 28.00 & 17.11 & 15.78 & 2.00 & 11.78 & 0.22 & 24.22 & 22.67 & 20.00 & 1.89 & 4.22 & 4.89 & 18.15 \\
MemoryVLA & 10.00 & 17.33 & 1.33 & 0.00 & 16.00 & 8.00 & 4.67 & 4.67 & 7.33 & 0.00 & 13.33 & 4.00 & 14.00 & 0.00 & 12.00 & 1.33 & 7.12 \\
SAM2Act+$^{\dagger}$ & 40.00 & 76.00 & 25.33 & 0.00 & 27.33 & 32.00 & 18.00 & 26.67 & 17.33 & 5.33 & 24.67 & 20.00 & 29.33 & 0.00 & 0.00 & 0.00 & 21.37 \\
T$^2$Mem & 60.67 & 91.33 & 90.67 & 70.00 & \textbf{88.00} & 59.33 & 29.33 & 20.00 & \textbf{50.67} & \underline{38.67} & 35.33 & 30.00 & 80.67 & \textbf{36.00} & 50.00 & \textbf{78.67} & 56.83 \\
\textbf{LTM-}\(\boldsymbol{\pi}\)\textbf{ (CD-LAM)} & \underline{64.67} & \textbf{100.00} & 93.33 & \underline{87.33} & 68.67 & \textbf{98.00} & \underline{30.67} & \underline{31.33} & 22.67 & 30.00 & 38.67 & 22.67 & 80.00 & \underline{20.67} & \textbf{64.00} & \underline{70.00} & \underline{57.67} \\
\textbf{LTM-}\(\boldsymbol{\pi}\)\textbf{ (DreamDojo)} & \textbf{66.00} & \underline{99.33} & \textbf{96.00} & \textbf{100.00} & \underline{72.67} & \underline{86.67} & \textbf{34.00} & \textbf{36.00} & \underline{38.67} & \textbf{42.67} & 38.00 & 30.67 & 77.33 & 20.00 & \underline{60.00} & 38.67 & \textbf{58.54} \\
\bottomrule
\end{tabular}%
}
\vspace{2pt}
\begin{minipage}{\textwidth}
\fontsize{7}{8}\selectfont
BF: BinFill; PX: PickXtimes; SX: SwingXtimes; SC: StopCube;
VU/BU: Video/ButtonUnmask; VUS/BUS: Video/ButtonUnmaskSwap;
PH: PickHighlight; VR: VideoRepick; VPB/VPO: VideoPlaceButton/Order;
MC: MoveCube; IP: InsertPeg; PL: PatternLock; RS: RouteStick.
$^{\dagger}$SAM2Act+ uses an external motion planner, but not an external reasoning model; its keyframe actions are control targets rather than memory-specific semantic annotations. An integrated VLM backbone is treated as part of a single policy, whereas a separate inference-time planner or reasoner is treated as an external reasoning model.
\end{minipage}
\end{table*}

As shown in Table~\ref{tab:robomme_results}, LTM-$\pi$ achieves average success rates of 57.67\% with CD-LAM and 58.54\% with DreamDojo, improving over $\pi_{0.5}$ by 39.74 and 40.61 percentage points, respectively. Within the block that uses neither external reasoning-model calls nor memory-specific annotations, the DreamDojo and CD-LAM variants rank first and second overall; DreamDojo exceeds the strongest baseline in this block, T$^2$Mem (56.83\%), by 1.71 percentage points. Both variants also outperform baselines, suggesting that compact transition representations provide an effective alternative to sampled-frame memory. The DreamDojo variant is particularly strong on Counting, achieving 90.33\% on average and solving StopCube with 100\% success. For context, SimpleMemVLA reaches 88.25\% using cloud-VLM-generated sub-task annotations, while PonderPounce reaches 60.83\% using a separate 9B context model and additional grounding and demonstration-reasoning supervision. LTM-$\pi$ achieves its results without external reasoning-model calls or memory-specific annotations. Reference tasks remain a target for improving fine-grained object identity and visual correspondence. In terms of inference efficiency, the average latency of a LTM-$\pi$ policy inference on an H100 is 66.71\,ms. Figure~\ref{fig:teaser} compares task-level success rates.

\begin{table*}[h]
\centering
\caption{Success rates (SR, \%) on LIBERO-Mem. Bold indicates the best result in each column.}
\label{tab:libero_mem_results}
\setlength{\tabcolsep}{3.5pt}
\renewcommand{\arraystretch}{1.12}

\resizebox{0.8\textwidth}{!}{%
\begin{tabular}{l*{11}{r}}
\toprule
Method
& T0 & T1 & T2 & T3 & T4
& T5 & T6 & T7 & T8 & T9
& Avg. \\
\midrule

$\pi_{0.5}$ (original)
& 0.0 & 0.0 & 0.0 & 0.0 & 0.0
& 0.0 & 0.0 & 0.0 & 0.0 & 0.0
& 0.0 \\

SlotVLA ($h=8$)
& 0.0 & 0.0 & 0.0 & 0.0 & 0.0
& 0.0 & 0.0 & 0.0 & 0.0 & 0.0
& 0.0 \\

Naive E-SlotSSM (oracle)
& 0.0 & 0.0 & 0.0 & 0.0 & 0.0
& 0.0 & 0.0 & 0.0 & 0.0 & 0.0
& 0.0 \\

\midrule

MemoryVLA
& 55.0 & 0.0 & 35.0 & 0.0 & 5.0
& 0.0 & 0.0 & 0.0 & 0.0 & 0.0
& 9.5 \\

GMP
& 85.0 & 45.0 & 20.0 & 5.0 & 40.0
& 0.0 & 0.0 & 0.0 & 0.0 & 0.0
& 19.5 \\

MemoryVAM (Pathway 1 only)
& 85.0 & 40.0 & 35.0 & 10.0 & 15.0
& 0.0 & 5.0 & 0.0 & 30.0 & 65.0
& 28.5 \\

MemoryVAM (Pathway 2 only)
& 45.0 & 10.0 & 5.0 & 0.0 & 5.0
& 5.0 & 0.0 & 0.0 & 5.0 & 0.0
& 7.5 \\

MemoryVAM (UNet)
& 85.0 & 65.0 & 90.0 & 10.0 & 35.0
& 0.0 & 25.0 & 5.0 & 40.0 & 70.0
& 42.5 \\

MemoryVAM (DiT)
& 75.0 & 65.0 & 65.0 & 15.0 & 55.0
& 35.0 & 10.0 & 10.0 & 25.0 & 50.0
& 40.5 \\

$\pi_0$ (2AM reimplementation)$^{\dagger}$
& 30.83 & 13.33 & 14.17 & 21.67 & 11.67
& 10.00 & 6.67 & 8.33 & 0.00 & 5.83
& 12.25 \\

2AM$^{\dagger}$
& 1.67 & 5.00 & 5.00 & 6.67 & 6.67
& 2.50 & 7.50 & 10.00 & 47.50 & 25.83
& 11.83 \\

\midrule

\multicolumn{12}{l}{\textbf{LTM-$\pi$ (Ours)}} \\
CD-LAM
& \textbf{95.0} & \textbf{75.0} & 90.0 & 70.0 & 70.0
& 45.0 & \textbf{80.0} & 20.0 & \textbf{100.0} & \textbf{100.0}
& 74.5 \\

DreamDojo
& \textbf{95.0} & \textbf{75.0} & \textbf{100.0} & \textbf{90.0} & \textbf{90.0}
& \textbf{65.0} & 65.0 & \textbf{25.0} & \textbf{100.0} & \textbf{100.0}
& \textbf{80.5} \\

\bottomrule
\end{tabular}%
}

\vspace{2pt}
\begin{minipage}{\textwidth}
\fontsize{7}{8}\selectfont
\raggedright
T0: Bowl back on plate; T1: Bottle back on plate;
T2: Bowl $\times 3$; T3: Bottle $\times 3$;
T4: Bowl $\times 5$; T5: Bowl $\times 7$;
T6: Swap 2 bowls; T7: Rotate 3 bowls;
T8: Cheese + basket center; T9: Cheese + empty basket center.
\par\vspace{2pt}
$^{\dagger}$ Their $\pi_0$ reimplementation uses Qwen3-VL-4B;
the rollout count is unspecified in the cited report, so a matched evaluation budget cannot be established.
\end{minipage}
\end{table*}

\paragraph{Results on LIBERO-Mem.}
As shown in Table~\ref{tab:libero_mem_results}, LTM-$\pi$ achieves average success rates of 74.5\% with CD-LAM and 80.5\% with DreamDojo, exceeding the strongest baseline included in this comparison, MemoryVAM (UNet), by 32.0 and 38.0 percentage points, respectively. DreamDojo improves over CD-LAM by 6.0 points overall, with particularly large gains on repeated-interaction tasks T3--T5, while both variants achieve 100\% success on T8 and T9. These results support the effectiveness of latent-transition memory for history-dependent manipulation across different LAM encoders. The encoder choice nevertheless affects individual tasks: CD-LAM performs better on swapping two bowls (T6), achieving 80.0\% versus 65.0\%. Rotating three bowls (T7) remains challenging for both variants (20.0\% and 25.0\%), suggesting that tracking multiple objects through sequential interactions remains a limitation.

\subsubsection{Ablations}
\label{sec:ablations}
We first isolate the contributions of transition content and visual anchors, then test how delaying memory admission affects success and online computation.

\paragraph{Effect of Latent Transitions and Visual Anchors.}
In Table~\ref{tab:robomme_ablation}, we ablate the two sources of historical information in LTM-$\pi$: latent-transition memory and sparse visual anchors. We compare the full model with \textbf{LT-only}, which removes visual anchors; \textbf{Anchor-only}, which removes latent transitions; and \textbf{Fixed Random LT}, which replaces observation-dependent latent transitions with fixed random representations while preserving the temporal structure of the memory sequence.

\begin{table*}[h]
\centering
\fontsize{8}{9}\selectfont
\setlength{\tabcolsep}{3pt}
\renewcommand{\arraystretch}{1.0}
\caption{Ablation study on the effect of latent transitions and visual anchors on RoboMME. Success rates (\%, $\uparrow$) are reported on all 16 tasks. Fixed Random LT removes transition-specific semantics while retaining temporal information from the memory sequence.}
\label{tab:robomme_ablation}
\resizebox{\textwidth}{!}{%
\begin{tabular}{@{}l@{\hspace{8pt}}*{4}{r}@{\hspace{9pt}}*{4}{r}@{\hspace{9pt}}*{4}{r}@{\hspace{9pt}}*{4}{r}@{\hspace{9pt}}r@{}}
\toprule
Method
& \multicolumn{4}{c}{Counting}
& \multicolumn{4}{c}{Permanence}
& \multicolumn{4}{c}{Reference}
& \multicolumn{4}{c}{Imitation}
& Avg. \\
\cmidrule(lr){2-5}
\cmidrule(lr){6-9}
\cmidrule(lr){10-13}
\cmidrule(lr){14-17}
& BF & PX & SX & SC
& VU & BU & VUS & BUS
& PH & VR & VPB & VPO
& MC & IP & PL & RS
& \\
\midrule
Full
& 64.67 & 100.00 & 93.33 & 87.33
& 68.67 & 98.00 & 30.67 & 31.33
& 22.67 & 30.00 & 38.67 & 22.67
& 80.00 & 20.67 & 64.00 & 70.00
& \textbf{57.67} \\

LT-only
& 66.00 & 100.00 & 94.00 & 90.00
& 24.00 & 44.00 & 30.00 & 16.00
& 26.00 & 30.00 & 40.00 & 36.00
& 90.00 & 14.00 & 46.00 & 40.00
& 49.13 \\

Anchor-only
& 34.00 & 32.00 & 42.00 & 0.00
& 66.00 & 58.00 & 24.00 & 8.00
& 12.00 & 2.00 & 26.00 & 10.00
& 32.00 & 2.00 & 12.00 & 10.00
& 23.13 \\

Fixed Random LT
& 72.00 & 74.00 & 98.00 & 58.00
& 82.00 & 94.00 & 30.00 & 38.00
& 18.00 & 26.00 & 26.00 & 20.00
& 60.00 & 2.00 & 10.00 & 4.00
& 44.50 \\
\bottomrule
\end{tabular}%
}
\end{table*}

Removing latent transitions reduces average success from 57.67\% to 23.13\%, while LT-only retains 49.13\%, showing that the transition sequence is the stronger historical component and anchors mainly provide complementary visual grounding.

Fixed Random LT still reaches 44.50\%, indicating that the ordered memory sequence itself provides useful cues about elapsed time and trajectory progress. Replacing these random tokens with observation-dependent latent transitions further improves performance to 57.67\%, demonstrating that transition-specific content contributes beyond temporal cues. This gain is limited on swap-dependent tasks, motivating the diagnostic study below.

\begin{table*}[h]
\centering
\fontsize{8}{9}\selectfont
\setlength{\tabcolsep}{3pt}
\renewcommand{\arraystretch}{1.0}
\caption{
Ablation study on lag $k$ on RoboMME.
Success rates (\%, $\uparrow$) are reported on all 16 tasks.
Reported LAM blocking is an estimated critical-path cost,
$\max(0,T_{\mathrm{LAM}}-T_{\mathrm{overlap}})$.
}
\label{tab:robomme_ablation_2}
\resizebox{\textwidth}{!}{%
\begin{tabular}{
@{}l@{\hspace{8pt}}
*{4}{r}@{\hspace{9pt}}
*{4}{r}@{\hspace{9pt}}
*{4}{r}@{\hspace{9pt}}
*{4}{r}@{\hspace{9pt}}
r@{\hspace{9pt}}
*{4}{r}@{}
}
\toprule
$k$
& \multicolumn{4}{c}{Counting}
& \multicolumn{4}{c}{Permanence}
& \multicolumn{4}{c}{Reference}
& \multicolumn{4}{c}{Imitation}
& Avg.
& \multicolumn{4}{c}{Runtime (ms, $\downarrow$)} \\
\cmidrule(lr){2-5}
\cmidrule(lr){6-9}
\cmidrule(lr){10-13}
\cmidrule(lr){14-17}
\cmidrule(lr){19-22}
& BF & PX & SX & SC
& VU & BU & VUS & BUS
& PH & VR & VPB & VPO
& MC & IP & PL & RS
& & Policy & Sim. & LAM update & LAM block \\
\midrule
0
& 62.00 & 98.00 & 94.00 & 90.00
& 70.00 & 94.00 & 30.00 & 34.00
& 28.00 & 26.00 & 38.00 & 24.00
& 82.00 & 20.00 & 66.00 & 70.00
& 57.88
& 66.71 & 117.01 & 114.25 (10p, sync.) & 114.25 \\
5
& 64.67 & 100.00 & 93.33 & 87.33
& 68.67 & 98.00 & 30.67 & 31.33
& 22.67 & 30.00 & 38.67 & 22.67
& 80.00 & 20.67 & 64.00 & 70.00
& 57.67
& 66.71 & 117.01 & 67.97 (5p, async.) & 9.47 \\
10
& 58.00 & 98.00 & 90.00 & 86.00
& 68.00 & 96.00 & 28.00 & 32.00
& 20.00 & 30.00 & 36.00 & 24.00
& 76.00 & 24.00 & 64.00 & 72.00
& 56.38
& 66.71 & 117.01 & 114.25 (10p, async.) & 0.00 \\
\bottomrule
\end{tabular}%
}
\end{table*}

\paragraph{Effect of Lag $k$.}
On RoboMME, we replan every 10 execution steps and use the same $k$ for training and inference. As shown in Table~\ref{tab:robomme_ablation_2}, lag creates computation slack between observing a transition and admitting it to policy memory. Increasing $k$ from 0 to 5 changes average success only from 57.88\% to 57.67\%, while reducing the estimated LAM blocking cost from 114.25\,ms to 9.47\,ms. At $k=10$, the measured LAM update fits within the available execution overlap, giving an estimated zero blocking cost, with 56.38\% success. These point estimates indicate that a small non-zero lag can reduce blocking with a small change in success. The 66.71\,ms policy time excludes LAM work; the blocking column is an overlap estimate, not a measured end-to-end control latency.

\subsubsection{Data Scaling on RoboMME}
\label{sec:data_scaling}
The ablations support the contribution of transition content beyond temporal cues. We next ask whether broader demonstration coverage helps the policy use this information across diverse task families. RoboMME provides 100 episodes per task family, each spanning multiple scenes and language instructions. Following the data-expansion setting explored by PonderPounce~\citep{choi2026ponderpounce}, we use RoboMME's generation pipeline to increase this to 1,000 episodes per family, yielding RoboMME-10x. Training-generation seeds are kept disjoint from validation and test seeds. We increase training from 40K to 120K steps and use a fixed lag $k=10$ during both training and inference. To assess whether longer training alone explains the improvement, we also train on the original data for 120K steps under the same configuration.

\begin{table*}[h]
\centering
\fontsize{8}{9}\selectfont
\setlength{\tabcolsep}{2.8pt}
\renewcommand{\arraystretch}{1.05}
\caption{Task-level data-scaling results on RoboMME (success rate, \%, $\uparrow$). We report PonderPounce (9B) with $1\times$ and $9\times$ data and LTM-$\pi$ with $1\times$ and $10\times$ data. Gain ratio is the increase in average success rate divided by the additional data multiple relative to $1\times$, measured in percentage points per additional $1\times$ data. The data, training, and supervision budgets differ between methods; gain ratio is descriptive, not a controlled estimate of sample efficiency.}
\label{tab:robomme_scaling}
\resizebox{0.98\textwidth}{!}{%
\begin{tabular}{@{}l@{\hspace{6pt}}c@{\hspace{6pt}}*{4}{r}@{\hspace{8pt}}*{4}{r}@{\hspace{8pt}}*{4}{r}@{\hspace{8pt}}*{4}{r}@{\hspace{8pt}}r@{\hspace{8pt}}r@{}}
\toprule
Method & Data scale & \multicolumn{4}{c}{Counting} & \multicolumn{4}{c}{Permanence} & \multicolumn{4}{c}{Reference} & \multicolumn{4}{c}{Imitation} & Avg. & Gain ratio \\
\cmidrule(lr){3-6}\cmidrule(lr){7-10}\cmidrule(lr){11-14}\cmidrule(lr){15-18}
& & BF & PX & SX & SC & VU & BU & VUS & BUS & PH & VR & VPB & VPO & MC & IP & PL & RS & & (pp/$\times$) \\
\midrule
PonderPounce (9B) & $1\times$ & 50.67 & 93.33 & 72.00 & 82.67 & 94.67 & 97.33 & 25.33 & 34.00 & 85.33 & 40.67 & 83.33 & 79.33 & 81.33 & 6.00 & 12.67 & 34.67 & 60.83 & -- \\
PonderPounce (9B) & $9\times$ & 74.00 & 98.00 & 93.33 & 60.00 & 96.67 & 100.00 & 73.33 & 50.67 & 92.67 & 88.67 & 92.00 & 97.33 & 93.33 & 18.67 & 25.33 & 54.67 & 75.54 & 1.84 \\
\addlinespace
LTM-$\pi$ (ours) & $1\times$ & 64.67 & 100.00 & 93.33 & 87.33 & 68.67 & 98.00 & 30.67 & 31.33 & 22.67 & 30.00 & 38.67 & 22.67 & 80.00 & 20.67 & 64.00 & 70.00 & 57.67 & -- \\
LTM-$\pi$ (ours) & $10\times$ & 80.00 & 100.00 & 100.00 & 100.00 & 100.00 & 100.00 & 74.00 & 62.00 & 70.00 & 66.00 & 84.00 & 84.00 & 96.00 & 88.00 & 98.00 & 98.00 & 87.50 & 3.31 \\
\bottomrule
\end{tabular}%
}
\end{table*}

Table~\ref{tab:robomme_scaling} shows that LTM-$\pi$ improves from 57.67\% to 87.50\%, a gain of 29.83 percentage points. Counting, Permanence, Reference, and Imitation rise from 86.33\%, 57.17\%, 28.50\%, and 58.67\% to 95\%, 84\%, 76\%, and 95\%, respectively. Reference gains the most (+47.50 points), showing that the weakest base-data track benefits substantially from the expanded training set. The scaled model reaches at least 88\% on every Imitation task and 100\% on three of four Counting tasks. At the original $1\times$ data scale, extending training from 40K to 120K steps yields no appreciable improvement. This control indicates that longer training alone does not account for the gain at $10\times$ data, supporting the contribution of additional demonstrations.

For comparison, PonderPounce improves from 60.83\% to 75.54\% with $9\times$ data (+14.71 points). At $10\times$ data, LTM-$\pi$ is 11.96 points higher overall, leading on 11 tasks and tying on ButtonUnmask; PonderPounce remains higher on the four Reference tasks. The descriptive gain ratios are 3.31 versus 1.84 percentage points per additional data multiple, respectively, under different training and supervision budgets. LTM-$\pi$ also comes within 0.75 points of SimpleMemVLA's reported 88.25\%, without memory-specific annotations but using more demonstrations (Table~\ref{tab:robomme_results}).

Despite the broad gains, VideoUnmaskSwap and ButtonUnmaskSwap reach only 74\% and 62\%, compared with 100\% on both non-swap counterparts. Expanded demonstration coverage therefore reduces, but does not remove, the difficulty of tracking state through swaps. We therefore turn to a controlled diagnostic to test whether the required swap information can be recovered from the transition sequence (Section~\ref{sec:swap_diagnostic}).

\subsubsection{Diagnostic Study: Can Latent Transitions Represent ``Swap''?}
\label{sec:swap_diagnostic}
\paragraph{Problem.}
The scaling results show that broader demonstration coverage improves overall performance, yet swap-dependent state tracking remains difficult. At $10\times$ data, the non-swap Video/ButtonUnmask tasks reach 100\%, while their swap counterparts reach 74\% and 62\% (Table~\ref{tab:robomme_scaling}). Our qualitative inspection also finds swap-related failures in VideoRepick. These results motivate a narrower question: \emph{is information needed to track swaps recoverable from the latent sequence under controlled conditions?}

\paragraph{Cup Swap Test.}
\begin{figure}[h]
  \centering
  \includegraphics[width=0.99\linewidth]{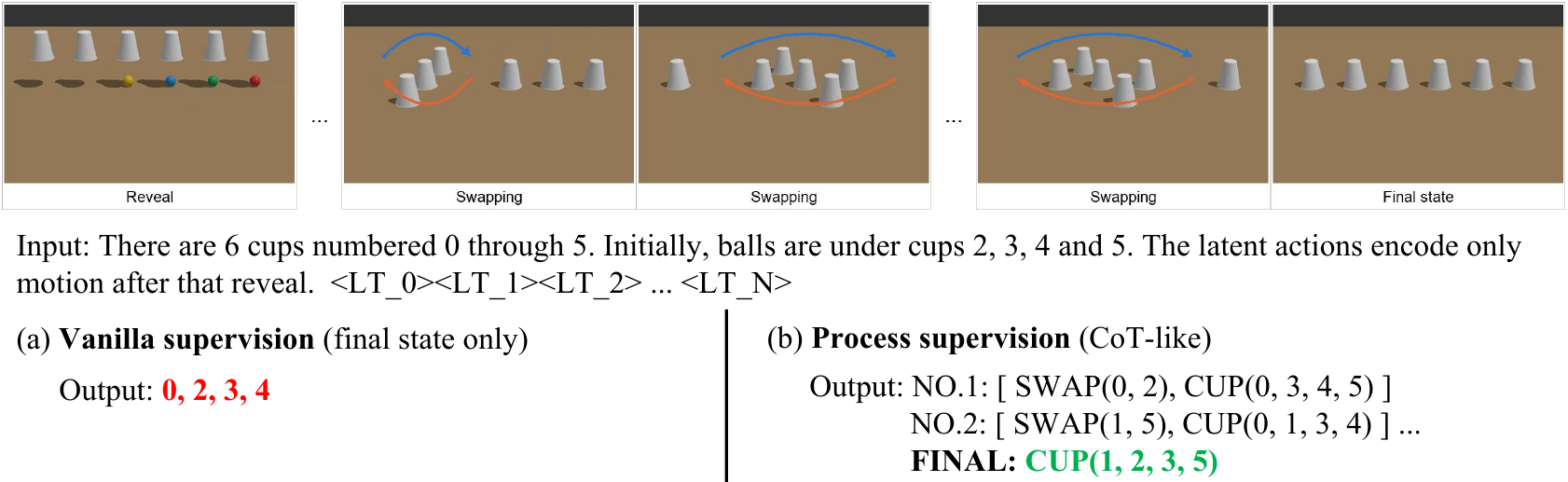}
  \caption{Our Cup Swap diagnostic benchmark and two supervision methods. The arrows are for illustration only and are not included in the actual dataset. \texttt{<LT\_i>} denotes a latent transition; the illustration uses zero-indexed latent labels.}
  \label{fig:swap}
\end{figure}
To test recoverability separately from robotic control, we construct the Cup Swap diagnostic shown in Figure~\ref{fig:swap}. Each sequence contains six cups and one to five hidden balls, followed by 5--20 randomly generated cup swaps. The initial ball locations are provided explicitly, while only the post-reveal motion history is represented by latent transitions. We inject these transitions into a Qwen3-VL-2B backbone and compare two supervision schemes: \textbf{Vanilla}, which directly predicts the final cup occupancy without intermediate supervision, and \textbf{Process supervision (CoT-like)}, which explicitly supervises swap tracking and the resulting intermediate/final state. The initial state is supplied as text, and this diagnostic uses a separate latent-fusion architecture rather than the LTM-$\pi$ policy. Details are provided in Appendix~\ref{app:swap_diagnostic}.

As shown in Table~\ref{tab:swap_diagnostic}, the process-supervised model recognizes individual swaps with 99.78\% accuracy and recovers final states with 97.93\% accuracy using a frozen VLM backbone, while the final-state-only model reaches 2.40\% best exact accuracy. These results demonstrate that latent transitions contain recoverable multi-step swap information and that structured intermediate supervision enables accurate temporal-state recovery in the diagnostic. The final-state-only model also has low training accuracy (approximately 12.5\%; Appendix~\ref{app:swap_diagnostic}), so optimization differences may contribute to the observed gap. The findings motivate process-guided temporal reasoning in future transition-memory control policies.

\begin{table}[h]
\centering
\caption{Diagnostic results (\%) on the six-cup, multi-ball Cup Swap task. Final-state accuracy requires exact cup occupancy. The Vanilla value is its best reported checkpoint result; its final checkpoint reaches 1.75\%.}
\label{tab:swap_diagnostic}
\begin{tabular}{lcc}
\toprule
Supervision & Individual Swap Acc. & Final-state Acc. \\
\midrule
Vanilla supervision (final state only) & -- & 2.40 \\
Process supervision (CoT-like) & 99.78 & 97.93 \\
\bottomrule
\end{tabular}
\end{table}

\Needspace{6\baselineskip}
\subsubsection{Token Compression: Can Transition Memory Remain Effective Under a Fixed Budget?}
\label{sec:token_compression}
The diagnostic demonstrates that process-supervised decoding recovers multi-swap state from the latent-transition sequence. We now return to LTM-$\pi$ control and ask whether a bounded context can retain enough information for effective manipulation. Although each transition occupies only one token, append-only execution memory still grows with episode duration. We evaluate DreamDojo-based LTM-$\pi$ on RoboMME-10x with a quota of $Q=200$ execution-transition tokens and a fixed lag $k=10$, using the same quota and compression rule during training and inference. The quota applies to execution transitions only; language, current visual inputs, and any fixed demonstration context are outside this count.

\paragraph{Online adjacent-token merging.}
Let $\mathcal U=(u_1,\ldots,u_n)$ denote the ordered list of retained tokens in the projector output space. When a new eligible transition token $m=P(\tilde z)$ arrives, we append it to $\mathcal U$. If the length exceeds $Q$, we select the most similar adjacent pair and replace it with its arithmetic mean:
\[
j^*=\underset{1\leq j<n}{\arg\max}\;\operatorname{cos}(u_j,u_{j+1}),
\qquad
\bar u=\frac{u_{j^*}+u_{j^*+1}}{2}.
\]
Here, $n$ is the length after appending, and $\operatorname{cos}$ is cosine similarity. The merged token occupies the earlier position and the later token is removed. We repeat this operation until the budget is satisfied (Algorithm~\ref{alg:token_compression}). Restricting merges to adjacent tokens preserves the order of the remaining temporal segments, while similarity guides compression toward locally redundant representations. The rule operates after projection and before the policy backbone, without an additional learned compressor. During training, each eligible history prefix is constructed by replaying these updates in chronological order; inference applies the same updates as transitions become eligible under the lag.

\begin{algorithm}[htbp]
\caption{Online Adjacent-Token Compression}
\label{alg:token_compression}
\begin{algorithmic}[1]
\Require Ordered projector-output tokens $\mathcal U$ with $|\mathcal U|\leq Q$, new projected token $m$, quota $Q\geq1$
\State $\mathcal U\gets\mathcal U\mathbin{\|}[m]$ \Comment{append in temporal order}
\While{$|\mathcal U|>Q$}
  \For{$j=1,\ldots,|\mathcal U|-1$}
    \State $s_j\gets\operatorname{cos}(u_j,u_{j+1})$
  \EndFor
  \State $j^*\gets\arg\max_j s_j$
  \State $u_{j^*}\gets(u_{j^*}+u_{j^*+1})/2$
  \State Delete $u_{j^*+1}$ from $\mathcal U$
\EndWhile
\State \Return $\mathcal U$
\end{algorithmic}
\end{algorithm}

Each update preserves a representative of every temporal segment instead of discarding the oldest tokens. However, the merge is lossy: it removes an event boundary and replaces two vectors with one. Repeated merges use equal weights for the two current representatives, without tracking their constituent transition counts, so the result is generally not the uniform mean of all original transitions in a segment. For a stream that is already within budget, appending one token triggers at most one merge; a direct scan costs $\mathcal O(Qd)$ per update for projector dimension $d$, and retained projected-token storage is $\mathcal O(Qd)$.

\Needspace{13\baselineskip}
\begin{wraptable}{r}{0.50\textwidth}
\vspace{-1pt}
\centering
\captionsetup{font=footnotesize,skip=5pt}
\caption{Execution-transition tokens per query ($k=10$). Uncompressed values are measured; compressed counts follow $\min(N,200)$. The compressed mean is an upper bound.}
\label{tab:token_budget}
\small
\setlength{\tabcolsep}{3pt}
\renewcommand{\arraystretch}{1.05}
\begin{tabular*}{\linewidth}{@{\extracolsep{\fill}}lrr@{}}
\toprule
Statistic & Uncompressed & \makecell[r]{Compressed\\($Q=200$)} \\
\midrule
Mean & 323.93 & $\leq200$ \\
Median & 265 & 200 \\
P90 & 747 & 200 \\
P95 & 893 & 200 \\
Maximum & 1,789 & 200 \\
Queries $>200$ (\%) & 61.98 & 0.00 \\
\bottomrule
\end{tabular*}
\vspace{-6pt}
\end{wraptable}
\paragraph{Token budget.}
We profile uncompressed execution memory at policy queries under the execution-only, fixed-$k=10$ training convention. The mean is 323.93 transition tokens/query, the median is 265, and the P90/P95 values are 747/893; the maximum is 1,789. In total, 61.98\% of queries exceed 200 tokens. After applying the rule, a prefix containing $N$ eligible transitions has exactly $\min(N,Q)$ retained tokens. Table~\ref{tab:token_budget} gives the resulting bounds for the same query population. The compressed mean is at most 200, implying a reduction of at least $1-200/323.93=38.26\%$ in mean execution-memory length; its exact value requires the full count distribution. At the uncompressed P95 and maximum, the counts fall from 893 and 1,789 to 200, reductions of 77.60\% and 88.82\%, respectively. These are token-count reductions, rather than measured inference speedups.

\paragraph{Manipulation results.}
Table~\ref{tab:token_compression} compares the compressed model with the uncompressed RoboMME-10x reference from Table~\ref{tab:robomme_scaling}. Compression achieves 672 successes over 800 rollouts, or 84.00\% average success, compared with 87.50\% without compression. This is a decrease of 3.50 percentage points, retaining 96.00\% of the reference success rate. Counting, Permanence, Reference, and Imitation average 91.50\%, 80.00\%, 74.00\%, and 90.50\%, respectively, versus 95.00\%, 84.00\%, 76.00\%, and 95.00\% without compression.

\begin{table*}[htbp]
\centering
\fontsize{8}{9}\selectfont
\setlength{\tabcolsep}{3pt}
\renewcommand{\arraystretch}{1.05}
\caption{DreamDojo-based LTM-$\pi$ on RoboMME-10x with and without execution-token compression (success rate, \%). The uncompressed row reproduces the $10\times$ reference in Table~\ref{tab:robomme_scaling}; the compressed model uses $Q=200$ during training and inference and is evaluated over 50 rollouts per task. $\Delta$ is compressed minus uncompressed success in percentage points. Task abbreviations follow Table~\ref{tab:robomme_results}.}
\label{tab:token_compression}
\resizebox{\textwidth}{!}{%
\begin{tabular}{@{}l@{\hspace{8pt}}*{4}{r}@{\hspace{9pt}}*{4}{r}@{\hspace{9pt}}*{4}{r}@{\hspace{9pt}}*{4}{r}@{\hspace{9pt}}r@{}}
\toprule
Memory & \multicolumn{4}{c}{Counting} & \multicolumn{4}{c}{Permanence} & \multicolumn{4}{c}{Reference} & \multicolumn{4}{c}{Imitation} & Avg. \\
\cmidrule(lr){2-5}\cmidrule(lr){6-9}\cmidrule(lr){10-13}\cmidrule(lr){14-17}
& BF & PX & SX & SC & VU & BU & VUS & BUS & PH & VR & VPB & VPO & MC & IP & PL & RS & \\
\midrule
Uncompressed & 80.00 & 100.00 & 100.00 & 100.00 & 100.00 & 100.00 & 74.00 & 62.00 & 70.00 & 66.00 & 84.00 & 84.00 & 96.00 & 88.00 & 98.00 & 98.00 & 87.50 \\
Compressed ($Q=200$) & 84.00 & 100.00 & 100.00 & 82.00 & 100.00 & 100.00 & 62.00 & 58.00 & 74.00 & 56.00 & 86.00 & 80.00 & 94.00 & 76.00 & 96.00 & 96.00 & 84.00 \\
$\Delta$ (pp) & +4.00 & 0.00 & 0.00 & -18.00 & 0.00 & 0.00 & -12.00 & -4.00 & +4.00 & -10.00 & +2.00 & -4.00 & -2.00 & -12.00 & -2.00 & -2.00 & -3.50 \\
\bottomrule
\end{tabular}%
}
\end{table*}

PickXtimes, SwingXtimes, VideoUnmask, and ButtonUnmask remain at 100\%, indicating that a bounded context can support these tasks despite substantial compression. The largest decreases occur on StopCube ($-18$ points), VideoUnmaskSwap and InsertPeg ($-12$ each), and VideoRepick ($-10$). This pattern is consistent with sensitivity to precise temporal boundaries or interaction details that averaging may weaken. In particular, the two swap tasks decline while their non-swap counterparts remain saturated. Cosine similarity measures representation similarity, rather than task relevance, so a locally similar pair can still contain an important state change. BinFill and PickHighlight improve by 4 points and VideoPlaceButton by 2, although single-seed evaluation does not resolve these smaller differences. Overall, a 200-token budget retains 96.00\% of the uncompressed success rate. Together with the swap diagnostic, they motivate preserving informative event boundaries through event-aware merging or explicit segment-duration information.

\paragraph{Compatibility with KV caching.}
The fixed token budget bounds execution-memory length independently of episode duration, making persistent KV caching less critical for computational efficiency. Although merging modifies existing prefix tokens and invalidates subsequent cached entries under causal attention, the bounded context allows these entries to be recomputed at a limited cost. Therefore, our compression strategy prioritizes bounded memory and context length over KV-cache reuse, which remains an optional optimization rather than a necessity.

\Needspace{7\baselineskip}
\subsection{Real-World Evaluation}
\label{sec:realworld_eval}
\subsubsection{Settings}
\paragraph{Platform.}
We use the Galbot G1 humanoid robot as the physical platform for real-world evaluation. During all experiments, only the two arms and grippers are actuated, while the remaining joints and body components are kept fixed.

\begin{figure}[h]
  \centering
  \includegraphics[width=0.99\linewidth]{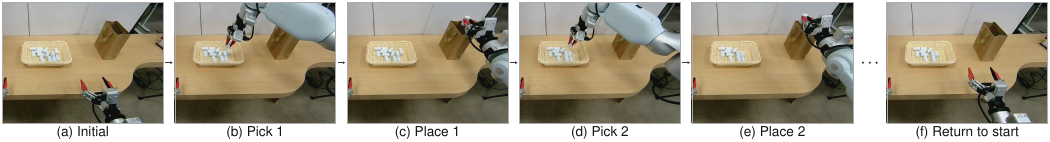}
  \caption{Task and environment for real-world experiments.}
  \label{fig:realtask}
\end{figure}

\paragraph{Task and Data.}
We consider a \emph{pick-and-place \(x\) times} task shown in Figure~\ref{fig:realtask}, where the robot is required to sequentially pick up \(x\) objects and place them into a packaging bag. We collect teleoperated demonstrations for \(x\in\{1,2,3,4\}\), with 50 trajectories for each setting, resulting in 200 trajectories in total. The dataset contains 222,834 frames recorded at 30 FPS, corresponding to 2.06 hours of real-world data.

\paragraph{Implementation Details.}
We run inference on a single RTX 4090 GPU. The robot is controlled with a control interval of \(\Delta t=0.05\,\mathrm{s}\) per step. For real-world evaluation, we do not use anchor frames. A trial is considered successful only if the robot transfers exactly the instructed number of objects, returns to its initial pose, and remains stationary for at least 10\,s.

\subsubsection{Results}

\begin{table}[h]
\centering
\small

\caption{Real-world success rate (\%, $\uparrow$) for picking and placing $x$ objects. Avg. is the unweighted mean across the four target quantities.}
\label{tab:realworld_results}
\begin{tabular}{lrrrrr}
\toprule
Method & $x=1$ & $x=2$ & $x=3$ & $x=4$ & Avg. \\
\midrule
$\pi_{0.5}$ (no memory) & 5.0 & 20.0 & 20.0 & 0.0 & 11.25 \\
LTM-$\pi$ (ours) & \textbf{95.0} & \textbf{90.0} & \textbf{70.0} & \textbf{75.0} & \textbf{82.50} \\
\bottomrule
\end{tabular}
\end{table}

Table~\ref{tab:realworld_results} shows that LTM-$\pi$ improves success rate across all four target quantities, achieving a success rate of 82.50\% compared with 11.25\% for the memory-free $\pi_{0.5}$, a gain of 71.25 percentage points. Qualitatively, the memory-free policy tends to continue picking until approximately $N$ objects remain in the source box, where $N$ is the mean remaining-object count observed during data collection, suggesting reliance on a familiar scene configuration rather than tracking completed transfers. By retaining interaction history, LTM-$\pi$ more reliably executes the requested number of transfers and stops accordingly. Per-trial transferred-object counts and a link to representative real-robot videos are provided in Appendix~\ref{app:realworld_raw}.

\paragraph{Additional Findings.}
We also observe shorter motion-onset delays for LTM-$\pi$ than for the memory-free policy in trials with long static segments at the start of demonstrations. This qualitative pattern is consistent with temporal-progress cues from the memory sequence; onset latency was not measured separately.

\section{Limitation}
Our method relies on a pretrained latent action model (LAM) to provide transition representations, making memory quality dependent on the information preserved by the encoder. While latent transitions effectively capture local interaction dynamics, they may not adequately represent substantial changes in scene content, such as the appearance of new objects, where visual appearance and semantic information become important. Although sparse visual anchors provide complementary information, they may not fully address this limitation. A promising direction is to combine latent transition memory with visual memory, exploring how to effectively integrate dynamic changes and visual context for more robust long-horizon manipulation.
\section{Conclusion}
We introduced Latent Transition Memory, which retains observed interactions as an ordered sequence of compact transition tokens. LTM-$\pi$ combines frozen latent action encoders with a lightweight adapter and cache-compatible causal attention, learning from action demonstrations without memory-specific semantic annotations or an external reasoning model during control. With DreamDojo transitions, it reaches 58.54\% success on RoboMME and 80.5\% on LIBERO-Mem at the original data scale. Ablations support the value of observation-dependent transition content beyond temporal-progress cues, with sparse visual anchors providing complementary grounding. Delayed memory admission also reduces estimated encoder blocking while largely preserving success.

We then examined the effect of demonstration scaling. Increasing RoboMME data tenfold and training from 40K to 120K steps raises success from 57.67\% to 87.50\%, with gains across all four tracks and the largest improvement on Reference. Extending training on the original data to 120K steps yields no appreciable gain, indicating that the improvement is not explained by longer training alone. These results support the benefit of broader demonstration coverage for learning to use compact transition memory.

Swap-dependent state tracking remains challenging after data scaling. The Cup Swap diagnostic shows that process-supervised decoding recovers final states from latent transitions with 97.93\% accuracy, compared with 2.40\% under final-state-only supervision. This demonstrates the recoverability of complex interaction information from transition memory and highlights intermediate supervision as a promising direction for temporal composition.

We next addressed execution-memory growth. Merging similar adjacent projected tokens with the same rule during training and inference caps execution memory at 200 tokens. This retains 84.00\% success on RoboMME-10x versus 87.50\% without compression, a decrease of 3.50 percentage points, while reducing mean transition-token count by at least 38.26\% on profiled training queries. These results demonstrate a trade-off between bounded context and manipulation performance. Cache refresh costs must also be considered when translating token-count reductions into runtime gains.

Finally, real-world experiments on a repeated pick-and-place counting task improve average success from 11.25\% with memory-free $\pi_{0.5}$ to 82.50\% with LTM-$\pi$. Together, these results establish compact latent transitions as an effective form of episodic memory for the evaluated history-dependent manipulation tasks. Improving temporal composition and preserving informative event boundaries are promising directions for further development.

\setlength{\bibsep}{3pt}
\bibliography{references}
\bibliographystyle{\colabbibstyle}

\clearpage
\appendix
\section{Cup Swap Diagnostic Experiment Details}
\label{app:swap_diagnostic}

\paragraph{Task and dataset.}
We construct a synthetic Cup Swap task to isolate motion-memory and temporal state-tracking ability from robotic control. Each video contains six cups and between one and five balls, with the initial balls placed under distinct cups. The initial configuration is revealed at frame 42, after which the sequence contains 5--20 randomly generated swaps. Videos are rendered at 30 FPS. The dataset contains 14,000 videos, split into 11,176 training, 1,375 validation, and 1,449 test examples using a fixed random seed of 42. The task is to recover the final ball occupancy after composing the complete swap sequence.

\paragraph{Latent-transition extraction.}
We use the same frozen encoder family as in LTM-$\pi$. Each adjacent frame pair is encoded into a 32-dimensional latent transition with stride 1. To prevent the reveal event itself from leaking the initial state into the motion representation, we retain only transitions whose source frame is at or after the reveal frame. Letting $r=42$ denote the reveal-frame index, we keep a transition $z_i$ only when its source-frame index satisfies $\operatorname{src}(z_i)\geq r$. Thus, the latent stream excludes the reveal transition; the initial occupancy is supplied separately as text.

\paragraph{Input and latent-memory fusion.}
To isolate motion reasoning from visual recognition of the reveal frame, the initial cup occupancy is provided as text (\texttt{reveal\_mode=text}). The base model is Qwen3-VL-2B-Instruct. Every four consecutive latents are compressed into one memory token and directly projected to the Qwen hidden dimension. We add sinusoidal temporal embeddings, process the latent stream with 8 layers of latent self-attention (16 heads), and let the language states read the latent memory through 8 layers of cross-attention (16 heads). Attention dropout is set to zero. Training uses bfloat16 and gradient checkpointing.

\paragraph{Supervision variants.}
We distinguish two objectives. \textbf{Vanilla (final-state only)} uses a cup-occupancy classification target and directly predicts whether each of the six cups contains a ball; it receives no supervision on the swap sequence or intermediate states. \textbf{Process supervision (CoT-like)} uses a process-query-tracking objective that supervises recovery of the swap process and the evolving state before predicting the final state. Although this objective does not generate a natural-language chain of thought, it explicitly supervises intermediate reasoning and is therefore CoT-like in function.

\paragraph{Training configuration.}
For the process-supervised run, we train on 8 H100 80GB GPUs with a per-GPU batch size of 4 (global batch size 32). The main learning rate is $1\times10^{-4}$, the classification-head learning rate is $3\times10^{-4}$, weight decay is $1\times10^{-2}$, and gradient clipping is 1.0. The Qwen backbone is frozen. 

\paragraph{Additional metrics.}
For the process-supervised model, token accuracy is 99.67\%, swap accuracy is 99.78\%, swap-count accuracy is 100.00\%, final-state accuracy is 97.93\%, and full exact accuracy is 97.52\%. For the Vanilla six-cup objective, the best exact accuracy is 2.40\% (1.75\% at the final checkpoint), with training exact accuracy of approximately 12.5\%. A separate, substantially simpler three-cup, single-ball, fixed-five-swap classifier reaches 99.40\% exact accuracy without process supervision, but we do not compare it directly because its state space and task difficulty are much smaller.

\Needspace{24\baselineskip}
\section{Real-World Evaluation}
\label{app:realworld_raw}
\paragraph{Raw Results.}
We report the actual number of objects transferred into the packaging bag in every trial in Table~\ref{tab:realworld_raw}. A trial is counted as correct if this number exactly matches the instructed quantity $x$, and the robot subsequently returns to its initial pose and remains stationary for at least 10\,s. Each model is evaluated for 20 trials per target quantity, except the memory-free $\pi_{0.5}$ at $x=3$ and $x=4$: we conducted only 10 trials each because the policy repeatedly failed to stop after the requested number of transfers and continued picking, making individual trials excessively long. Unperformed trials are marked as ``--'' and excluded from the accuracy denominator.

\begin{table}[htbp]
\centering
\small
\setlength{\tabcolsep}{6pt}
\renewcommand{\arraystretch}{1.0}
\caption{Per-trial transferred-object counts in real-world evaluation. Bold entries match the instructed quantity; ``--'' denotes an unperformed trial. Trial indices are local to each model--quantity setting.}
\label{tab:realworld_raw}
\begin{tabular}{lrrrrrrrr}
\toprule
& \multicolumn{4}{c}{$\pi_{0.5}$ (no memory)} & \multicolumn{4}{c}{LTM-$\pi$ (ours)} \\
\cmidrule(lr){2-5}\cmidrule(lr){6-9}
Trial & $x=1$ & $x=2$ & $x=3$ & $x=4$ & $x=1$ & $x=2$ & $x=3$ & $x=4$ \\
\midrule
1 & 0 & 1 & 4 & 5 & \textbf{1} & \textbf{2} & \textbf{3} & \textbf{4} \\
2 & 2 & 1 & 5 & 7 & \textbf{1} & \textbf{2} & \textbf{3} & 3 \\
3 & 2 & \textbf{2} & 5 & 8 & \textbf{1} & 1 & \textbf{3} & 3 \\
4 & 2 & 3 & 4 & 6 & \textbf{1} & \textbf{2} & 4 & \textbf{4} \\
5 & 3 & 3 & 6 & 8 & \textbf{1} & \textbf{2} & 4 & \textbf{4} \\
6 & 4 & 3 & \textbf{3} & 5 & \textbf{1} & \textbf{2} & \textbf{3} & \textbf{4} \\
7 & 4 & 5 & 4 & 7 & \textbf{1} & \textbf{2} & \textbf{3} & \textbf{4} \\
8 & 3 & 4 & \textbf{3} & 6 & \textbf{1} & \textbf{2} & 2 & \textbf{4} \\
9 & 3 & 3 & 6 & 6 & 2 & \textbf{2} & \textbf{3} & 3 \\
10 & 4 & 3 & 4 & 8 & \textbf{1} & \textbf{2} & \textbf{3} & 3 \\
11 & 3 & \textbf{2} & -- & -- & \textbf{1} & \textbf{2} & \textbf{3} & \textbf{4} \\
12 & 2 & 3 & -- & -- & \textbf{1} & \textbf{2} & \textbf{3} & \textbf{4} \\
13 & 3 & 1 & -- & -- & \textbf{1} & \textbf{2} & 2 & \textbf{4} \\
14 & \textbf{1} & 1 & -- & -- & \textbf{1} & \textbf{2} & 4 & \textbf{4} \\
15 & 2 & \textbf{2} & -- & -- & \textbf{1} & \textbf{2} & \textbf{3} & \textbf{4} \\
16 & 3 & 3 & -- & -- & \textbf{1} & \textbf{2} & \textbf{3} & \textbf{4} \\
17 & 3 & 3 & -- & -- & \textbf{1} & \textbf{2} & \textbf{3} & \textbf{4} \\
18 & 5 & 4 & -- & -- & \textbf{1} & \textbf{2} & \textbf{3} & \textbf{4} \\
19 & 4 & \textbf{2} & -- & -- & \textbf{1} & 1 & \textbf{3} & 3 \\
20 & 3 & 3 & -- & -- & \textbf{1} & \textbf{2} & 4 & \textbf{4} \\
\midrule
Correct / Trials & 1/20 & 4/20 & 2/10 & 0/10 & 19/20 & 18/20 & 14/20 & 15/20 \\
Acc. (\%) & 5.0 & 20.0 & 20.0 & 0.0 & 95.0 & 90.0 & 70.0 & 75.0 \\
\bottomrule
\end{tabular}
\end{table}

\paragraph{Representative Real-Robot Videos.}
Representative videos of our real-robot experiments, including successful LTM-$\pi$ pick-and-place trials and a representative failure of the memory-free $\pi_{0.5}$ baseline, are available in our project repository: \url{https://github.com/buaa-colalab/LTM-pi}.

\end{document}